\documentclass[a4paper]{article}

\usepackage[sort]{cite}
\usepackage{listings}
\usepackage{longtable}
\usepackage{booktabs}

\usepackage{tikz}
\usepackage{graphicx}

\usepackage{tipa}

\usepackage{hyperref}

\usepackage{draftwatermark}
\SetWatermarkText{DRAFT}
\SetWatermarkScale{5}

\usepackage{url}
\urldef{\mymail}\path|tresoldi@gmail.com|

\begin{document}

\title{A tentative model for dimensionless phoneme distance from binary
distinctive features}
\author{
Tiago Tresoldi \\
Universidade Federal do Rio Grande (FURG) \\
\mymail
}
\maketitle

\begin{abstract}
This work proposes a tentative model for the calculation of
dimensionless distances between phonemes; sounds are described with
binary distinctive features and distances show linear consistency in
terms of such features. The model can be used as a scoring function for
local and global pairwise alignment of phoneme sequences, and the
distances can be used as prior probabilities for Bayesian analyses on
the phylogenetic relationship between languages, particularly for
cognate identification in cases where no empirical prior probability is
available. \\
\textbf{Keywords}: phonetic similarity, cognancy detection, string comparison.
\end{abstract}

\section{Introduction}\label{introduction}

Since the work on lexicostatistics by Morris Swadesh in the 1950s there
has been an increasing interest in the usage of quantitative methods for
research in historical linguistics. This interest has increased
significantly in the last decades due to the expansion of computing
power, the availability of structured digitalized linguistic data, and
the adoption of methods first developed for biological research, such,
as remembered by List (2012), phylogenetic analysis (\emph{e.g.}, Ringe,
Warnow, and Taylor (2002); Gray and Atkinson (2003); Holman et al.
(2011)) and statistical aspects of genetic relationship (\emph{e.g.},
Baxter and Ramer (n.d.); Mortarino (2009)). The new methods have drawn
substantial criticism from more traditionally oriented researchers, as
exemplified by the review of Gray and Atkinson (2003) by both H{\"a}kkinen
(2012) and Pereltsvaig and Lewis (2015). Among the most common
objections, which tend to extend the criticisms to Joseph Greenberg's ``mass comparison'' (such as in Campbell (2001)), is a general opposition to analyses that rely
exclusively in lexical data, where the difficulty in controlling for
false cognates and borrowed lexemes is contrasted by the regularity of
standard reconstructions focused in morphological and phonological
elements.

When considering these criticisms, the obvious alternative is to replace
or extend cognate sets of lexical data from the comparison of the
surface level of candidates for cognancy in terms of phonetic
similarity. However, phonetic similarity is a very broad term without an
accepted or even possible definition, often, as stated by Mielke (2012),
``invoked for explaining a wide range of phonological observations'':
while the notion of ``phoneme'' as a distinctive unit of sound in a
language is not disputed, the differentiation of actual phonemes across
languages, particularly in the case of historical analysis with inferred
data, is contextual and depends on the level of detail agreed upon when
describing segments of sound. In short, a single and universal model for
phoneme similarity is impossible. To face this limitation, most models
of phonetic similarity don't offer a formal description of their scoring
functions, but are guidelines used by researchers and their tools when
deciding the most likely candidates for a sound change or
correspondence, usually based in similarities of place and manner of
articulation and in the frequency of sounds, sound changes, and sound
correspondences. While quantitative models for well-studied families
such as Indo-European can be derived from observable data, in other cases, particularly when
the relationship itself is the object of investigation, the models are
forcibly based in data extrapolated from known families and intuition.

As in the case of Mielke (2012), this work offers a model for
``quantifying phonetic similarity, {[}\ldots{}{]} and for distinguishing
phonetic similarity from phonological notions of similarity, such as
those based on features {[}\ldots{}{]} or on phonological patterning''.
It intends to offer dimensionless distances between phonological
segments described with a broad set of distinctive features; in
particular, there are no separate submodels based in phonological
classes, allowing researchers to compare between any two segments
(including between vowels and consonants). Despite its many limitations,
some of which are described in the last section, by ultimately relying
in acoustic and articulatory properties, only complemented by perceptual
ones when necessary, this model can be used as a reference for future
models or as a source of prior probabilities for Bayesian analyses,
particularly in cases for which no model of phonetic or phonological
similarity can be computed from observable data or agreed upon by
experts.

\subsection{Review of alternative
models}\label{review-of-alternative-models}

As in the review by List (2012) of automatic approaches for cognancy
detection in multilingual word lists, whose description of the process
is here condensed, most methods analyze the surface similarity of
phonetic sequences, such as words and morphemes, by calculating a
``phonetic distance''. This analysis usually builds upon the paradigm of
sequence alignment, where the phonemes of the sequences are arranged in
a matrix in such a way that corresponding phonemes occupy the same
column, with empty cells filled with gap symbols resulting from
non-corresponding segments (Gusfield (1997), 216). Each matched
\emph{residue pair} is given a specific phoneme similarity score based
on the \emph{scoring function} at the core of the analysis. The
calculation of a normalized distance score from the individual distances
allows to determine cognancy, which is assumed in the case of a sequence
score under a certain threshold.

Scoring functions are essentially of two types: those that consider the
\emph{edit distance}, which only distinguish identical from
non-identical segments (thus returning the same score for a different
but related pair such as /p/ and /b/, and for a different and more
dissimilar pair such as /p/ and /a/), and those which return individual
scores based in the similarity of the phonemes being compared, such as
the ALINE algorithm of Kondrak (2002) and the sound-class models after
the work of Dolgopolsky (1964); the first type is marginally useful and
reserved for the comparison of very similar sequences, while the second
is recommended for complex, multi-language studies. According to List
(2012), the advantage of these methods is that the similarity of
phonetic segments is not determined on the basis of phonetic features,
but on the probability that their segments occur in a correspondence
relation in genetically related languages, as resulting by the
comparison of sequences with respect to their \emph{sound classes}. In
his original study based on an empirical analysis of sound
correspondence frequencies in a sample of over 400 languages,
Dolgopolsky (1964) proposed ten fundamental sound classes, with the idea
of ``{[}dividing{]} sounds into such groups, that changes within the
boundary of the groups {[}would be{]} more probable than transitions
from one group into another'' (List's translation from the Russian
original of Burlak and Starostin, 2005, 272), determining cognancy by
comparing the classes of the first two consonants of word roots.

There are known limitations to the model of Dolgopolsky, such as
sometimes failing to match attested cognates in the common case of sound
changes that operate at a suprasegmental level. An example given by List
(2012) is the correspondence between the German word \emph{Tochter}
\textipa{[tOxt@r]} and the English word \emph{daughter} \textipa{[dO:t@r]}, a false
negative in the model of Dolgopolsky that does not consider the
lengthening of the English vowel that precedes the cancelled consonant
as a correspondence to the German fricative. Alternatives include SCA,
an extension of Dolgopolsky's model by List (2012), and the Automated
Similarity Judgment Program (ASJP), an independent sound-class model
developed in the context of language classification by Holman et al.
(2011); these three models are used in LiNGPY (List and Forkel (2016)).
Bibliographic review has identified some additional proposals, usually
with \emph{ad hoc} individual scores established by researchers based on
personal experience and intuition; however, no actual numeric data for
these proposals is available to the general public.

An alternative to these phonological models, which as stated operate on
classes of observed historic sound changes and whose principle of
naturalness and applicability of sound changes to all languages is
debatable, are models of phonetic distance. Exemples are models that try
to quantify phoneme descriptions in terms of place and manner of
articulation, such as in calculations of the Euclidean distance between
vowels in the vowel trapezium, and those based in acoustic measurements,
such as the one developed by Mielke (2012) and here extended. However,
as far as bibliographic research has shown, this last model has never
been used for cognancy detection, an omission that likely results, at
least in part, from its reliance in distinctive features, the theoretic
fundamental units of phonological structure described as traits that
distinguish among natural classes of segments.

It is important to note that Mielke (2008) refutes the idea that
distinctive features have a biological basis, a property that is usually
assumed given their extended usage in generative theory and particularly
in Chomsky and Halle (1991), and adopts the different and explicitly
``non-natural'' theory of \emph{feature emergence}. In fact, while
supposedly ``natural'', the actual distinctive features have even less
consensus than standard descriptions in terms of place and manner of
articulation: the development of the theory can be traced back to
Nikolai Trubetzkoy's proposal of privative phonological oppositions, but
the more serious developments have been conducted by Roman Jakobson,
which served as a starting point for the more established models of
Chomsky \& Halle, Halle \& Clements and Ladefoged. Mielke (2008) states
that the motivation for his metric was the investigation of the role of
phonetic similarity in determining the sets of segments involved in
sound patterns, extending his earlier argument, in explicit contrast to
Chomsky \& Halle, that ``an all-purpose model is in conflict with many
of the attested phonologically active classes, including recurrent ones,
and that the preference for certain classes (e.g., vowels, nasals) is
driven by physiological and perceptual factors and their role in
diachronic change''.

As described by the author, his model is a metric for measuring phonetic
similarity based on several types of cross-linguistic phonetic data, in
particular inventory frequency. Data also came from the acoustic
production of segments by trained phoneticians and native speakers, with
both audio and video recording, from which numeric features of
production were extracted, yielding measurements such as oral airflow,
nasal airflow, vocal fold contact area, larynx height, acoustic
principal components, and vocal tract principal components. These values
were combined with measures of phonological similarity formulated using
the sound patterns reported in a database and software called
\emph{P-base}, which contained 6,077 phonologically active classes that
serve as the trigger or target for an alternation, many of which are
involved in multiple sound patterns within the same language. Discussing
the results, Mielke states that some familiar patterns ``are seen in the
phonetic and phonological similarity {[}\ldots{}{]}, involving the
patterning of non-sibilant voiced fricatives, glottals, and the
association of particular contrast with particular types of data'';
apparently strange behavior, such as the patterning of trills and flaps,
is attributed by the author to non disclosed ``methodological issues''.
Despite its limitation, particularly the number of covered phonemes,
this model was elected as the basis for the one here proposed.

\subsection{The new and tentative
model}\label{the-new-and-tentative-model}

The model here presented is part of a larger project for applying
quantitative methods to historical linguistics; among its goals are the
estimation of probability for ancestral states represented as phonemes.
The model was developed along the following guidelines:

\begin{itemize}
\item
  distances between phonemes are expressed in a dimensionless scale
  between 0.0, the score for the comparison of a phoneme with itself,
  and 1.0, the largest possible score in the model;
\item
  the model allows the comparison of different phonemes with a null
  state (i.e., no sound), allowing to quantify sound creation and
  cancellation;
\item
  there is a unique model for all phonemes, without separate models for
  separate sound classes;
\item
  while not considering disputed phonemes or phonemes with extremely
  marginal presence in inventories, the model covers all theoretically
  possible phonemes, including non-pulmonic and laryngeal consonants;
\item
  the model is based in a single set of binary distinctive features
  allowing individual phoneme comparisons that can be later extended to
  other models using more common descriptions, such as place and manner
  of articulation.
\end{itemize}

\subsubsection{Model development}\label{model-development}

The complete sequence of analysis rules and model development can be
obtained in a repository hosted on GitHub and in the related Python
notebook at \url{https://github.com/tresoldi/alterphono}. This section offers
a broad description of the obstacles found and solutions adopted.

Given the requirements for the model listed above, after investigating
which feature systems could be used as a basis for the one here
presented the decision rested on the feature set of \emph{Phoible} by
Moran, McCloy, and Wright (2014), a repository of cross-linguistic
phonological inventory data extracted from source documents and tertiary
databases. For every segment in its database, \emph{Phoible} includes a
distinctive feature description based on Hayes (2009) with additions
from S. R. Moisik and Esling (2011), and intended for adequate
cross-linguistic description. It should be stressed that the feature set
in \emph{Phoible} is descriptive and not exclusive as in other models,
such as in Chomsky \& Halle: for example, it includes both
\texttt{labial} and \texttt{labiodental} features, which in most other
systems are explained by combining different features or by reducing
them to a single feature. Future models will probably use a different
set of features, particularly the one from
Department~of~Linguistics~--~UCSB (2016).

The review of the bibliography by Mielke (2008) on distinctive features
suggested incorporating and extending his work on phoneme similarity
based on acoustic properties, as described in the previous section, by
combining the distinctive features used in \emph{Phoible} with the
phoneme comparison scores offered by Mielke (2012). Unfortunately,
neither the \emph{P-Data} phonotactic resources nor the phoneme distance
matrix were available at the address listed in Mielke (2008); an old
version of \emph{P-Data} was obtained from an archived page hosted at
University of Ottawa, not available at the time of writing, and a
partial list of phonetic similarity scores was obtained from Savva et
al. (2014). This matrix of phonetic similarity includes only a limited
number of phonemes (51), with around 1,300 scores; among its
limitations, it does not include a single entry for many distinctive
features used in \emph{Phoible}, such as \texttt{advancedTongueRoot} and
\texttt{long}, and does not include non-pulmonic consonants. The data
from this reduced Mielke's matrix was extended with phoneme and
allophone distribution data across language inventories, combining
resources from \emph{Phoible}, from \emph{Fonetikode} by Dediu and
Moisik (2015), and from Creanza et al. (2015), with minimal
normalization accounting for language relatedness. The results from the
first regression models were far from optimal, as the combination of
language variables (such as co-occurrence across inventories and
allophone distribution) and distinctive features properties resulted in
datasets that exceeded the limits for numeric generalization.

The set of distinctive features for all phonemes in the available Mielke
matrix was then combined with the scores from the same matrix; all
features from \emph{Phoible} were included, with the exception of
\texttt{stress} and \texttt{tone}. The first was not included given our
goals (suprasegmental qualities should be indicated at sequence, and not
at phoneme level) and the second was excluded for being used in
\emph{Phoible} to distinguish tone marks from phonemes, and not to
indicate the tonality of vowels. The dataset dimension was reduced by
converting the categorical features of \emph{Phoible} to logical
features; null values, which represent ``non applicable'', were
considered logical negatives. Scores were normalized in the range 0.0 to
1.0 and some manual corrections were performed in face of
inconsistencies in the IPA representation between the \emph{Phoible}
database and Mielke's matrix (for example, converting the stop in
palato-alveolar affricates in \emph{Phoible}, explicitly marked as
dentals, to alveolars).

As expected, the intercepts for the first linear models were around 1.0,
with mostly negative coefficients. Data exploration confirmed that
acoustic features in general had larger coefficients, so that manner of
articulation had higher scores than place of articulation. For example,
the single largest negative coefficient was the one for when both
phonemes are non strident, indicating a proximity for \texttt{-strident}
phonemes, while some features for place of articulation have positive
coefficients (such as \texttt{+back}, \texttt{+coronal} and
\texttt{+labiodental}), suggesting that the difference between phonemes
that share these traits are even more dependent on manner of
articulation. Even though neural networks are said to resemble black
boxes, models of different topologies were developed and studied to
understand the relationships, that were clearer given that the limited
amount of data inevitably resulted in over-fitting; the neuron
connections indicated that, while there is interaction between features,
it was not significant enough to justify a multidimensional and
non-linear model for our purposes.

Data exploration confirmed that, despite the strong multicollinearity
problems as expected from the strong feature correlation and the reduced
number of phoneme comparisons available in Mielke's matrix, it was
acceptable to base the model after linear regressions with parameters
estimated by Ordinary Least Squares (OLS). The first models performed
fairly enough in most comparisons of phonemes not included but similar
to those available in the matrix, a decision that was considered
consistent with phonetic theory. Among the main differences with the
scores of Mielke's model, this first model rated the mean distance
between stops and fricatives lower and with less standard deviation than
Mielke's model, whose model also tends to assign a larger weight to
vowel openness than vowel backness (mean distances of 0.35 and 0.19,
respectively), essentially equivalent in the model here proposed. At the
same time, this model returns higher mean distances than Mielke when
comparing stops and affricates, particularly in the case of different
places of articulation, and tends to return higher values than Mielke
when comparing vowels to voiced consonants. In general, these
differences are probably due to a dependence of Mielke's model in
acoustic values which seems to be computed on a case by case basis,
while our model is linear consistent.

\subsubsection{Solving problems with Mielke's
data}\label{solving-problems-with-mielkes-data}

As mentioned above, Mielke's matrix covers a small subset of the
segments listed in \emph{Phoible}, and in particular there are nine
distinctive features not covered by any segment:
\texttt{advanced\ tongue\ root}, \texttt{click},
\texttt{epilaryngeal\ source}, \texttt{fortis}, \texttt{long},
\texttt{lowered\ larynx}, \texttt{raised\ larynx},
\texttt{retracted\ tongue\ root}, \texttt{short}. As the intended model
is not a measure of acoustic and inventory data but a dimensionless
scale constructed from these data, acting as a proxy for quantifying the
differences perceived by expert phoneticians, the matrix needed to be
extended with inferences, including data points that would allow the
regression algorithms to generalize across the entire segment inventory.
The essential points for this inferences are described in this section
and the complete list of included data points can be found in the source
code.

\paragraph{Advanced and retracted tongue
root}\label{advanced-and-retracted-tongue-root}

\emph{Phoible} includes four advanced tongue root segments and 115 retracted tongue root segments (mostly pharyngealized
vowels and consonants). ATR and RTR, as commonly abbreviated, are
contrasting states of the root of the tongue in the production of sounds
(in most cases, vowels), particularly common in languages of West and
East Africa. Phonetically, ATR and RTR involve an expansion or a
contraction, respectively, of the pharyngeal cavity. In the case of ATR,
the larynx is lowered during the pronunciation, adding a breathy
quality; in the case of RTR, the retraction generally has an effect of
partial pharyngealization. Considering that almost every language that
presents these features also exhibit some kind of vowel harmonization
system that could be simulated from available data points, information
from the Fante dialect of the Akan language from Stewart (1967) was
combined with a provisional system for Brazilian Portuguese from Lee
(2013) to establish an ``advanced tongue root delta'' from the distance
of proximal vowels in terms of mouth opening and backness, which seems
reasonable considering other deltas based exclusively in place of
articulation.

While it would have been valid to replicate the calculation for tongue
root retraction, for example calculating the distance between voiced
uvular and epiglottal stops, /\textscg{}ɢ/ and /\textbarglotstop{}ʡ/, or between corresponding
fricatives, /\textinvscr{}ʁ/ and /\textipa{Q}ʕ/, Mielke's matrix unfortunately offers no
pharyngeal or epiglottal phoneme. The value for advanced tongue root
delta was thus replicated to retracted tongue root delta.

\paragraph{Non-pulmonic consonants}\label{non-pulmonic-consonants}

For modeling non-pulmonic consonants, clicks, ejectives, and implosives
(features \texttt{click}, \texttt{raisedLarynxEjective}, and
\texttt{lowered\-Larynx\-Implosive} in \emph{Phoible}) were considered
similar sounds related to stop consonants and differing exclusively in
manner of articulation. It was assumed that clicks can technically be
described as obstruents articulated with two points of mouth closure, in
which the enclosed pocket of air is rarefied by a sucking action of the
tongue; in acoustic terms, this assumes that, all other airflow
variables being equal, clicks are the loudest possible speech sounds.
Considering that one class is frequently mistaken for the other, and
that processes of allophony and sound change commonly involve both
classes, the modeling also considered that ejectives are the consonants
closer to clicks. Regarding implosives, it was assumed that, among
non-pulmonic consonants, they are the ones acoustically closer to
standard stops, as the implosion is caused by simply pulling the glottis
downwards, expanding the vocal tract.

Given these considerations, a central non-pulmonic delta was stipulated
as numerically equivalent to the largest possible difference in manner
of articulation, computed after a mean difference between stops and
corresponding affricates, for a value of circa 0.28 (some stop/affricate
pairs were excluded because the model performance was clearly
inadequate). A second delta of circa 0.24, stipulated as equivalent to
the mean value between corresponding stops and fricatives, and a third
delta, stipulated as half the mean distance between corresponding stops
and ejectives, were calculated to add data points relative to
non-pulmonic consonants to our dataset.

\paragraph{Fortis}\label{fortis}

Feature \texttt{fortis}, sometimes confused or combined with the feature
\texttt{voice}, is commonly used to distinguish between a standard and a
more ``pronounced'' articulation of the same sound, usually a consonant;
it tends to be used in contrast with \texttt{lenis}, a feature missing
in \emph{Phoible} mostly used to denote an underlying change in
pronunciation that can have many different surface forms, such as in
voicing/devoicing, aspiration/deaspiration, glottalization,
velarization, phoneme lengthening, lengthening of nearby vowels, etc. In
\emph{Phoible}, however, the feature \texttt{fortis} seems to be used
exclusively to indicate phonemes for languages where the contrast
between stops do not involve voicing. Considering that the feature is
probably superfluous for the goals of this model, as well as its low
number of occurrences, it was stipulated that \texttt{fortis} would be
the mean value between the voiced and the voiceless version of a
phoneme.

\paragraph{Long and short}\label{long-and-short}

As in the case of \texttt{fortis} and \texttt{lenis}, \texttt{long} is
usually considered an underlying, deep feature that has different
surface realizations, the most common being vowel lengthening for vowels
and gemination for consonants. Considering that this kind of description
frequently considers the difference between flaps and trills as an
expression of an underlying \texttt{long} feature, the mean value of
distance between corresponding flaps and trills in Mielke's matrix was
stipulated as equivalent to a ``long delta''.

\subsubsection{Normalization and
intuition}\label{normalization-and-intuition}

After the stipulation of the deltas described in the subsections above,
the models were run a successive number of times, manually adjusting
missing data points according to phonetic theory, to the author's
intuition and to discussions with colleagues. The complete list of
adjustments can be obtained in the related Python notebook.

\section{Distances}\label{distances}

The table below presents the scores for the comparison of the 10 most
common segments in \emph{Phoible}:

\begin{longtable}[c]{@{}lrrrrrrrrrr@{}}
\toprule
& /a/ & /i/ & /j/ & /k/ & /m/ & /n/ & /p/ & /s/ & /u/ & /w/\tabularnewline
\midrule
\endhead
/a/ & 0.00 & & & & & & & & &\tabularnewline
/i/ & 0.29 & 0.00 & & & & & & & &\tabularnewline
/j/ & 0.27 & 0.01 & 0.00 & & & & & & &\tabularnewline
/k/ & 0.38 & 0.31 & 0.32 & 0.00 & & & & & &\tabularnewline
/m/ & 0.25 & 0.25 & 0.23 & 0.41 & 0.00 & & & & &\tabularnewline
/n/ & 0.25 & 0.25 & 0.20 & 0.43 & 0.12 & 0.00 & & & &\tabularnewline
/p/ & 0.37 & 0.29 & 0.35 & 0.11 & 0.41 & 0.45 & 0.00 & & &\tabularnewline
/s/ & 0.89 & 0.77 & 0.80 & 0.64 & 0.99 & 0.99 & 0.69 & 0.00 &
&\tabularnewline
/u/ & 0.21 & 0.19 & 0.22 & 0.32 & 0.23 & 0.26 & 0.31 & 0.94 & 0.00
&\tabularnewline
/w/ & 0.24 & 0.22 & 0.22 & 0.34 & 0.19 & 0.24 & 0.29 & 0.97 & 0.01 &
0.00\tabularnewline
\bottomrule
\end{longtable}

The full dataset allows to perform a principal components analysis to
identify, by visual exploration or by some clustering method, the groups
of phonemes. In Figure 1, comprising the first two components, it is
possible to identify two main groups, roughly of continuant and
non-continuant sounds.

\begin{figure}[htbp]
\centering
\includegraphics[width=1.0\textwidth]{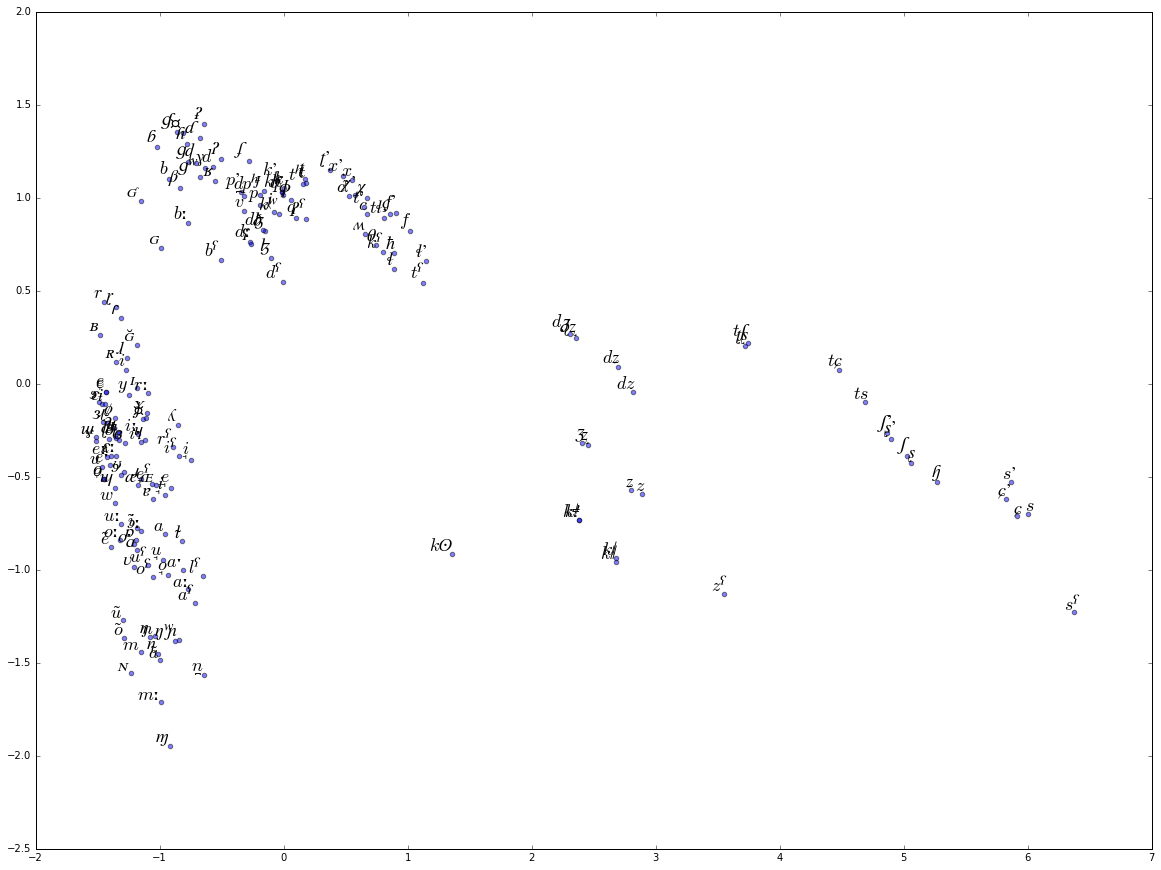}
\caption{Scatterplot with the first two components of the PCA
of our final matrix}
\end{figure}

\section{Alignment}\label{alignment}

Algorithms for word alignment and cognate detection, such as those used in LingPy, can be extended using the distances here presented. The tables below show the scores of the matrix here presented with a number of cognate sets included in the tests of LingPy, with a calculation that tries to emulate the results given by LingPy.

\subsection{Test 1 - ``naming who must not be named''}

From the popular book series, it is a set of words used in the tutorials of LingPy. Words: \textipa{woldemort}, \textipa{waldemar}, \textipa{wladimir}, \textipa{vladymir}.

\begin{table}[!htbp]
\centering
\caption{Cognancy score for Test \#1}
\label{table-test-1}
\begin{tabular}{lrrrr}
          & \textipa{woldemort} & \textipa{waldemar} & \textipa{wladimir} & \textipa{vladymyr} \\
\textipa{woldemort} & -         & +21.93   & -1.55    & -11.06   \\
\textipa{waldemar}  & +21.93    & -        & +4.30    & -4.71    \\
\textipa{wladimir}  & -1.55     & +4.30    & -        & +27.93   \\
\textipa{vladymyr}  & -11.06    & -4.71    & +27.93   & -       
\end{tabular}
\end{table}

\subsection{Test 2 - ``I (ego)''}

The first person pronoun. Words: \textipa{un@} (Albanian), \textipa{Z@} (French), \textipa{ix} (German), \textipa{Si} (Navajo), \textipa{ben} (Turkish). English and Hawaiian words were removed due to differences in the treatment of diphtongs in LingPy algorithms and in the matrix here presented, which should be solved in the future.

\begin{table}[!htbp]
\centering
\caption{Cognancy score for Test \#2}
\label{table-test-2}
\begin{tabular}{lrrrrr}
    & \textipa{un@}    & \textipa{Z@}     & \textipa{ix}     & \textipa{Si}     & \textipa{ben}    \\
\textipa{un@} & -      & -32.45 & -19.26 & -76.25 & -14.14 \\
\textipa{Z@}  & -32.45 & -      & -19.69 & -31.74 & -57.45 \\
\textipa{ix}  & -19.26 & -19.69 & -      & -15.00 & -31.30 \\
\textipa{Si}  & -76.25 & -31.74 & -15.00 & -      & -52.49 \\
\textipa{ben} & -14.14 & -57.45 & -31.30 & -52.49 & -     
\end{tabular}
\end{table}

\subsection{Test 3 - ``all''}

Synonym of "everything". Words: \textbardotlessj\textipa{iT} (Albanian), \textipa{Ol} (English), \textipa{tut} (French), \textipa{al} (German), \textipa{bytyn} (Turkish). Navajo and Hawaiian words were removed due to differences in the treatment of diphtongs in LingPy algorithms and in the matrix here presented, which should be solved in the future.

\begin{table}[!htbp]
\centering
\caption{Cognancy score for Test \#3}
\label{table-test-3}
\begin{tabular}{lrrrrr}
      & \textbardotlessj\textipa{iT}    & \textipa{Ol}      & \textipa{tut}    & \textipa{al}     & \textipa{bytyn}  \\
\textbardotlessj\textipa{iT}   & -      & -27.63  & -0.02  & -28.49 & -35.45 \\
\textipa{Ol}    & -27.63 & -       & -25.99 & +5.96  & -36.43 \\
\textipa{tut}   & -0.02  & -25.99  & -      & -27.67 & -27.88 \\
\textipa{al}    & -28.49 & +5.96   & -27.67 & -      & -38.11 \\
\textipa{bytyn} & -35.45 & -36.43 & -27.88 & -38.11 & -     
\end{tabular}
\end{table}

\subsection{Test 4 - ``animal''}

Synonym of "beast". Words: \textipa{kafS} (Albanian), \textipa{an@m@l} (English), \textipa{animal} (French), \textipa{ti:r} (German), \textipa{holoholona} (Hawaiian), \textipa{na:lde:hi} (Navajo), \textipa{hajvan} (Turkish).

\begin{table}[!htbp]
\centering
\caption{Cognancy score for Test \#4}
\label{table-test-4}
\begin{tabular}{lrrrrrrr}
           & \textipa{kafS}    & \textipa{an@m@l}  & \textipa{animal}  & \textipa{ti:r}   & \textipa{holoholona} & \textipa{na:lde:hi} & \textipa{hajvan}  \\
\textipa{kafS}       & -       & -75.00  & -75.00  & -79.44 & -166.47    & -107.30   & -57.79  \\
\textipa{an@m@l}     & -75.00  & -       & +24.10  & -90.00 & -109.01    & -58.54    & -70.00  \\
\textipa{animal}     & -75.00  & +24.10  & -       & -90.00 & -109.01    & -56.10    & -70.00  \\
\textipa{ti:r}       & -79.44  & -90.00  & -90.00  & -      & -93.23     & -49.11    & -90.00  \\
\textipa{holoholona} & -166.47 & -109.01 & -109.01 & -93.23 & -          & -88.97    & -76.57  \\
\textipa{na:lde:hi}  & -107.30 & -58.54  & -56.10  & -49.11 & -88.97     & -         & -105.00 \\
\textipa{hajvan}     & -57.79  & -70.00  & -70.00  & -90.00 & -76.57     & -105.00   & -      
\end{tabular}\end{table}

\subsection{Test 5 - ``earth''}

Synonym of "soil". Words: \textipa{Der} (Albanian), \textipa{@rT} (English), \textipa{tEr} (French), \textipa{e:rd@} (German), \textipa{lepo} (Hawaiian), \textbeltl\textipa{e:Z} (Navajo), \textipa{toprak} (Turkish).

\begin{table}[!htbp]
\centering
\caption{Cognancy score for Test \#5}
\label{table-test-5}
\begin{tabular}{lrrrrrrr}
       & \textipa{Der}    & \textipa{@rT}    & \textipa{tEr}    & \textipa{eːrd@}  & \textipa{lepo}   & \textbeltl\textipa{e:Z}   & \textipa{toprak} \\
\textipa{Der}    & -      & -12.38 & -11.03 & -21.84 & -27.16 & -15.65 & -34.94 \\
\textipa{@rT}    & -12.38 & -      & -11.49 & -11.50 & -38.19 & -43.58 & -29.95 \\
\textipa{tEr}    & -11.03 & -11.49 & -      & -21.32 & -46.03 & -36.32 & -23.66 \\
\textipa{e:rd@}  & -21.84 & -11.50 & -21.32 & -      & -14.49 & -43.73 & -31.53 \\
\textipa{lepo}   & -27.16 & -38.19 & -46.03 & -14.49 & -      & -35.70 & -39.37 \\
\textbeltl\textipa{e:Z}   & -15.65 & -43.58 & -36.32 & -43.73 & -35.70 & -      & -72.15 \\
\textipa{toprak} & -34.94 & -29.95 & -23.66 & -31.53 & -39.37 & -72.15 & -     
\end{tabular}
\end{table}

\section{Discussion and future work}\label{discussion-and-future-work}

According to a common aphorism in statistics attributed to George Box,
all models are wrong, but some are useful. The first part of the
aphorism is certainly applicable to the model here presented: apart from
the inescapable shortcomings of a universal model based on the abstract
concept of ``phoneme'', the final matrix has a number of limitations,
such as the omission of tones and suprasegmental features and an
inconsistent treatment of phoneme grouping (for example, affricates are
treated as a single unit, while diphthongs are not). An even more
consistent objection is that it consists of a one-rate model offered for
applications where two-rated models would be expected and needed. This,
in fact, is an essential limitation of the matrix here presented, as it
proposes a solution based in \emph{acoustic} and \emph{perceptual}
differences for the quantification of the \emph{transitional}
differences of historical linguistics, in which it is well established
that sound changes in a given direction are more natural and expected
than the inverse ones. For example, in many different languages (Ancient
Greek, Proto-Iranian, Caribbean Spanish, etc.), the evolution from /s/
to /h/, from an alveolar to a glottal fricative, is expected as a
consequence of the anticipatory widening of the glottis that allows an
adequate airflow for the voiceless fricative, while the sound change in
the opposite direction is not attested; at the same time, the
palatalisation of velars before front vowels is well established,
``whereas the backing and hardening of postalveolar or alveolar
fricatives before non-front vowels is virtually unheard of'' (both
examples are from Weiss (2015)).

The author believes, however, that the model can be useful for some
applications of language evolution modeling (including the generation of
random datasets for stressing algorithms). In particular, considering
attested sound changes and the Bayesian approach to this kind of method
that dominates the field, the scores should provide a useful prior
probability for hypothesis testing, possibly more helpful than
attributing the same probability to every transition or diving phonemes
in classes based in their vowelness, voiceness, or place of
articulation, as these categories, useful for synchronic description,
have limited usefulness in diachronic research; the distances provided
by these model should accelarate the estimation of parameters from
actual, historical data on language evolution. At last, the author
believes that the model can be used as a scoring function for local and
global pairwise alignment of phoneme sequences, both in global alignment
analyses based on the algorithm by Needleman and Wunsch (1970) and in
alignment analyses which maximize local similarities as in Morgenstern,
Dress, and Werner (1996).

Future work should extend this model with language specific data,
particularly for the research of the development of Indo-European
phonetic inventories. A new model with a set of distinctive features that extends and replaces the one used in Phoible is under development at 
\url{https://github.com/tresoldi/alterphono/tree/master/new_model}.

\bibliographystyle{IEEEtran}

\end{document}